\documentclass[letterpaper, 10 pt, conference]{ieeeconf}  % Comment this line out if you need a4paper

\IEEEoverridecommandlockouts                              % This command is only needed if 
\usepackage{epsfig} % for postscript graphics files
\usepackage{times} % assumes new font selection scheme installed
\usepackage{amsmath} % assumes amsmath package installed
\usepackage{amssymb}  % assumes amsmath package installed
\usepackage{graphicx}

\graphicspath{ {./images/},{./images/gta2cs/}, {./images/segmentation/}, {./images/segmentation_dl/}, {./images/synth2cs/}}

\title{\LARGE \bf
KLIEP-based Density Ratio Estimation for Semantically Consistent Synthetic to Real Images Adaptation in Urban Traffic Scenes
}

\author{Artem Savkin$^{1,2}$\\TUM, BMW \and Federico Tombari$^{1,3}$\\TUM, Google% <-this % stops a space
\thanks{$^{1}$TU Munich, Boltzmannstr. 3, 85748 Munich (Germany)
{\tt\small artem.savkin@tum.de};
{\tt\small tombari@in.tum.de}}%
\thanks{$^{2}$BMW AG, Petuelring 130, 80809 Munich (Germany)
}%
\thanks{$^{3}$Google, Brandschenkestrasse 110, 8002 Zurich (Switzerland)
}%
}

\begin{document}

\newcommand{\TODO}[1] {\textbf{[TODO: #1]}}
\newcommand{\Loss}{\mathcal{L}}
\newcommand{\Exp}{\mathop{\mathbb{E}}}
\newcommand{\rarr}{\rightarrow}
\newcommand{\larr}{\leftarrow}

\maketitle

\begin{abstract}
Synthetic data has been applied in many deep learning based computer vision tasks. Limited performance of algorithms trained solely on synthetic data has been approached with domain adaptation techniques such as the ones based on generative adversarial framework. We demonstrate how adversarial training alone can introduce semantic inconsistencies in translated images. To tackle this issue we propose density prematching strategy using KLIEP-based density ratio estimation procedure. Finally, we show that aforementioned strategy improves quality of translated images of underlying method and their usability for the semantic segmentation task in the context of autonomous driving.
\end{abstract}


\section{Introduction}
Transition of deep learning from being a mere research topic to application in a wide spectrum of industrial task made availability of comprehensive training data exceptionally crucial. Certain safety critical contexts additionally have particular requirements on reliability. In deep learning based computer vision common approach to achieve such a capacious training corpus would be just to acquire and label more data when it needs to cover any specific case. For autonomously driving systems that means to drive specific scenarios and improve models on newly captured data. However due to high costs (new scenarios should be driven and manually labeled), corner cases (are rare to capture) and near-accident scenarios (ethical issues) this strategy is not always fully applicable in autonomous driving.

In this regard synthetically generated data seem to be a natural solution to for the stated problem. And the straight-forward approach would be to utilize rendering engines to generate data which could be used in computer vision tasks. This not only could potentially extend the variability of training data at reduced cost but also minimize manual effort in labeling data. Thus many researchers focused on utilizing 3D rendered imagery in theirs approaches in computer vision tasks \cite{Taylor2007}.
\begin{figure}[!ht]
%\begin{center}
\begin{minipage}{0.495\columnwidth}
\includegraphics[width=1\textwidth]{02428_gta}
\includegraphics[width=1\textwidth]{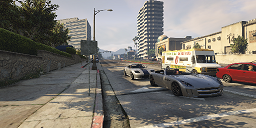}
\includegraphics[width=1\textwidth]{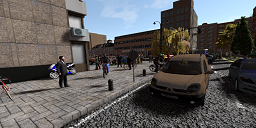}
\includegraphics[width=1\textwidth]{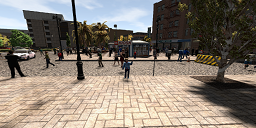}
\centering{Original synthetic}
\end{minipage}
\begin{minipage}{0.495\columnwidth}
\includegraphics[width=1\textwidth]{02428_cyclegan}
\includegraphics[width=1\textwidth]{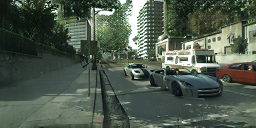}
\includegraphics[width=1\textwidth]{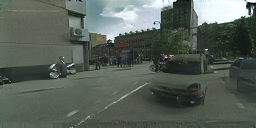}
\includegraphics[width=1\textwidth]{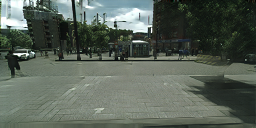}
\centering{Translated to real}
\end{minipage}
\caption{Example of semantical inconsistency introduced by adversarial training under covariate shift.}
%\end{center}
\label{fig:mismatch}
\end{figure}
Although rendered training data provides an opportunity to simulate various scenarios it reveals limited applicability in real-world environment. In machine learning one commonly considers training and validation data to be \textit{independent and identically distributed (iid)}. This is however clearly does not hold for synthetic-real setup, as even photo-realistically rendered images reveal bias on the underlying domain. Deep models trained solely on rendered images show poor performance when evaluated on real data \cite{Ros2016}.\\
This situation is commonly referred to as \textit{domain shift} and is considered to be the main reason for such performance. Particular case where input distribution for a model changes is referred to as \textit{covariate shift} and is addressed by means of \textit{domain adaptation}. Recent domain adaptation techniques enables to improve performance compared to models trained synthetically but still can not achieve same-domain results.

State-of-the-art domain adaptation methods such as DTN \cite{Taigman2016}, FCN ITW \cite{Hoffman2016} or DualGAN \cite{Yi2017} rely on \textit{generative adversarial network} \cite{Goodfellow2014}, which employs adversarial training for translating between source and target domains \cite{Hong2019}. During such training two networks generator and classifier (discriminator) perform a minimax game where the first one learns to conduct certain perturbations in the input samples from source domain $\{x_i^s\} \in D_s$ so that discriminator cannot distinguish them from the target domain samples $\{x_j^t\} \in D_t$. Thus GAN indirectly imposes target distribution upon the generated distribution \cite{Goodfellow2014}. Adversarial training being very efficient in adaptation tasks is a subject for covariate shift itself and it does not guarantee that a non-linear transformation performed by a generator keeps underlying semantical structure of the source inputs unchanged. Regularities in the target data learned by the discriminator are implicitly inflicted on generated samples.

Examples where adversarial network translates samples semantically inconsistent could be observed on the figure~\ref{fig:mismatch}. Here one can see vegetation patches imposed on sky regions or road users removed from the traffic scene. As seen on the figure~\ref{fig:mismatch} the network introduces semantically mismatching artifacts in order to reconstruct the target distribution. Such mutations in a semantic layout of the image reduce usability of generated data for computer vision tasks e.g. semantic segmentation or detection. Semantically inconsistent adaptation is especially critical in the area of traffic scenes understanding as it produces unreliable training data.

Multiple works investigated the ways to mitigate this problem and ensure that macro-structure of translated images remains consistent. They introduced dedicated constraints such as self-regularization loss \cite{Shrivastava2016} semantic consistency loss \cite{Zhu2017}, regularization by enforcing bijectivity \cite{Hoffman2016}, or modeling a shared latent space \cite{Liu2016}, \cite{Liu2017}, or semantic aware discriminator \cite{Li2018} to reduce undesired changes.

In this work we propose \textit{density ratio based distribution pre-matching} in ensemble with cyclic-consistency loss for adversarial synthetic to real domain adaptation in traffic urban scenes. For the density ratio estimation we employ Kullback-Leibler importance estimation procedure (KLIEP) \cite{Sugiyama2007}. This helps to keep  semantic consistency of translated images and improves visual quality of generated samples. Being evaluated on the particular task of semantic segmentation it reveals better average performance and performance for main classes. It does not affect the stability of adversarial training as it avoids additional constraints and losses.\\
\begin{figure*}[!h]
%\begin{center}

\begin{minipage}{0.195\textwidth}
\includegraphics[width=1\textwidth]{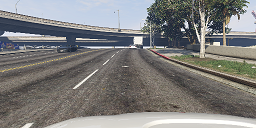}
\includegraphics[width=1\textwidth]{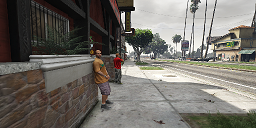}
\includegraphics[width=1\textwidth]{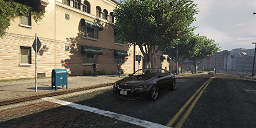}
\includegraphics[width=1\textwidth]{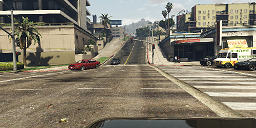}
\centering{No adaptation}
\end{minipage}
\begin{minipage}{0.195\textwidth}
\includegraphics[width=1\textwidth]{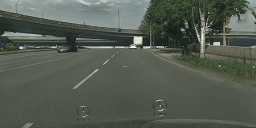}
\includegraphics[width=1\textwidth]{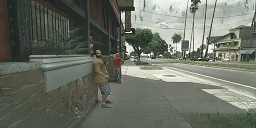}
\includegraphics[width=1\textwidth]{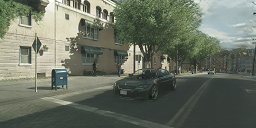}
\includegraphics[width=1\textwidth]{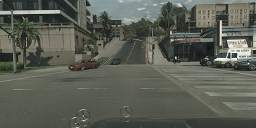}
\centering{CyCADA}
\end{minipage}
\begin{minipage}{0.195\textwidth}
\includegraphics[width=1\textwidth]{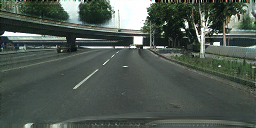}
\includegraphics[width=1\textwidth]{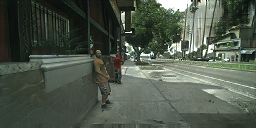}
\includegraphics[width=1\textwidth]{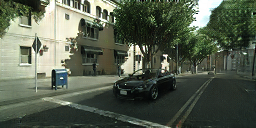}
\includegraphics[width=\textwidth]{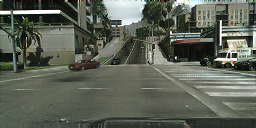}
\centering{CycleGAN}
\end{minipage}
\begin{minipage}{0.195\textwidth}
\includegraphics[width=1\textwidth]{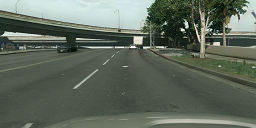}
\includegraphics[width=1\textwidth]{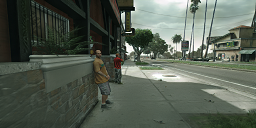}
\includegraphics[width=1\textwidth]{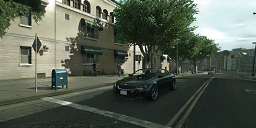}
\includegraphics[width=1\textwidth]{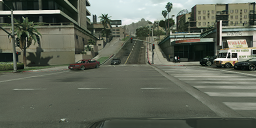}
\centering{UNIT}
\end{minipage}
\begin{minipage}{0.195\textwidth}
\includegraphics[width=1\textwidth]{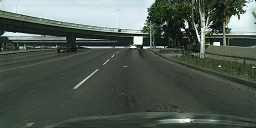}
\includegraphics[width=1\textwidth]{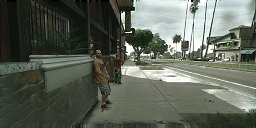}
\includegraphics[width=1\textwidth]{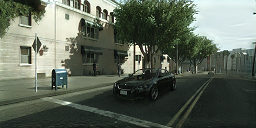}
\includegraphics[width=1\textwidth]{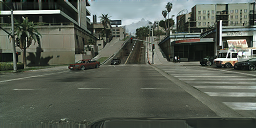}
\centering{Ours}
\end{minipage}

\caption{Examples of images adapted from synthetic to real.}
%\end{center}
\label{fig:adaptation}
\end{figure*}
\section{Related Work}
Synthetic data has found its application in variety of computer vision tasks. Hattori \textit{et al.} used spatial information of virtual scene to create surveillance detector \cite{Hattori2015}.

It also has been widely used for evaluation purposes. \cite{Kaneva2011} used virtual worlds to test feature descriptors and \cite{Handa2014} used synthetically generated environments for evaluation on such tasks as visual odometry or SLAM.

There are plethora of research works which utilized CAD models for computer vision tasks. %\cite{Stark2010} used 3D models for learning 2D shapes that can be matched with images, 
Sun and Saenko in \cite{Sun2014} investigated 3D models for 2D object detection and \cite{Aubry2014} to establish part based correspondences between 3D CAD models and real images.
\cite{Peng2015} showed effectiveness of augmentation of training data with crowd-sourced 3D models and \cite{Pepik2015} extended part models to include viewpoint and geometry information for joint object localization and viewpoint estimation.

Another vivid research area which utilizes rendered data is motion and pose estimation. For example, \cite{Shotton2013} used realistic and highly varied training set of synthetic images to learn model invariant to body shape, clothing and other factors. \cite{Varol2017} presented SURREAL large-scale dataset with realistically generated images from 3D human motion sequences.

Synthetically generated data seem to be especially useful when labeling of real data is tedious. This is the case with pixel dense tasks such as flow and depth estimation.  Dosovitskiy \textit{et al.} \cite{Dosovitskiy2015} generated unrealistic synthetic dataset called Flying Chairs and showed good generalization abilities of flow estimators. \cite{Handa2016} focused on depth-based semantic per pixel labeling an \cite{Papon2015} sets up a on-the-fly rendering pipeline to generate cluttered rooms for indoor scene understanding, which is also a subject of investigation in \cite{Satkin2012}

Considering almost eternal variability of traffic scenarios it is natural that synthetic data extended its area of application to traffic scenes understanding. In particular, pedestrian detection got a lot of attention. \cite{Marin2010}, \cite{Li2016}, \cite{Vazquez2014} addressed the question transfer learning trying to answer whether a pedestrian detector learned in virtual environment could work with real images.

There are also certain synthetic datasets for traffic scenes. \cite{Haltakov2013} provided accurate flow, depth and segmentation ground-truth for approximately 8,000 frames. Ros \textit{et al.} developed one of the major datasets in this area called SYNTHIA \cite{Ros2016}. Gaidon \textit{et al.} in \cite{Gaidon2016} introduced a virtual-to-real clone method to create so called "proxy virtual worlds" and released "Virtual KITTI" dataset. Synthetic dataset with the highest variance in scenes and scenarios counting almost 25,000 densely annotated frames provided in \cite{Richter2016}.

Severe disadvantage of those utilizing rendered data is that models trained on synthetic data generalize rather poor in real world. This issue as already mentioned is commonly known as \textit{domain shift} \cite{Sugiyama2012} and addressed by \textit{domain adaptation} techniques. Recent synthetic to real domain adaptation techniques could be roughly fall into 2 categories and both of them commonly rely on adversarial training. One category incorporates adversarial loss directly into task learning procedure. They commonly use both synthetic and real images as an input producing segmentation maps (or any other CV task output). Such models normally do not generate additional data. Although adversarial loss assist to bridge the gap between synthetic and real traffic scenes it is not cut for target accurate classification or detection learning task. Thus, multiple approaches introduce various regularization techniques. \cite{Saito2017} and \cite{Luo2019} utilizes discrepancy loss to generate target features close to source. \cite{Tsai2018} applies adversarial loss directly on learned segmentation features maps. Another examples are \cite{Sadat2018}, \cite{Wu2019}, \cite{Chen2018CVPR} and \cite{Zou2018}, \cite{Chen19}.

Another category of methods focuses on translation of synthetic images to real ones and using them afterwards for target prediction learning, they also are called generative. Here adversarial loss allows to generate visually pleasing images of high resolution by minimizing the distance between generated and target distributions. Many researchers focused their efforts to design dedicated constrains in adversarial models to overcome the mismatch problem. CycleGAN \cite{Zhu2017} uses cyclic-consistency in addition to adversarial loss. CyCADA \cite{Hoffman2017} improves on top of \cite{Zhu2017} by integrating the segmentation loss and \cite{Fu2019} introduced geometry consistency loss. Some works introduced disentanglement of content and appearance in a latent space \cite{Liu2017}, \cite{Shu2018}.

In our approach we focus on synthetic to real image transfer and handle domain shift issue by using the \textit{importance weighting} technique based on density ratio estimation. Certain works utilize importance weights \cite{Hjelm2017} or kernel density \cite{Sinn2017} to improve GAN training. We employ a technique named KLIEP \cite{Sugiyama2007} to pre-match distribution densities alongside with adversarial and cycle consistency loss, that allows us to perform semantically consistent synthetic to real domain adaptation in unsupervised manner. We show that our model shows significant performance improvement in data generation compared to state-of-the-art synthetic to real generative models (second category). This is evaluated qualitatively on on the task of semantic segmentation. Our ablation study shows how KLIEP based importance pre-matching affects adversarial training of our model.

\begin{table*}[!t]
%\begin{center}
\renewcommand{\arraystretch}{1.5}
\resizebox{\textwidth}{!}{
\begin{tabular}{l|cc|ccccccccccccccccccc}
\hline
\rotatebox[origin=c]{90}{ Method }
& \rotatebox[origin=c]{90}{ Accuracy }
& \rotatebox[origin=c]{90}{ mean IoU }
& \rotatebox[origin=c]{90}{ road }
& \rotatebox[origin=c]{90}{ sidewalk }
& \rotatebox[origin=c]{90}{ building }
& \rotatebox[origin=c]{90}{ wall }
& \rotatebox[origin=c]{90}{ fence }
& \rotatebox[origin=c]{90}{ pole }
& \rotatebox[origin=c]{90}{ traffic light }
& \rotatebox[origin=c]{90}{ traffic sign }
& \rotatebox[origin=c]{90}{ vegetation }
& \rotatebox[origin=c]{90}{ terrain }
& \rotatebox[origin=c]{90}{ sky }
& \rotatebox[origin=c]{90}{ person }
& \rotatebox[origin=c]{90}{ rider }
& \rotatebox[origin=c]{90}{ car }
& \rotatebox[origin=c]{90}{ truck }
& \rotatebox[origin=c]{90}{ bus }
& \rotatebox[origin=c]{90}{ train }
& \rotatebox[origin=c]{90}{ motorbike }
& \rotatebox[origin=c]{90}{ bicycle } \\
\hline

CS \cite{Cordts2016} & 94.3 & 67.4 & 97.3 & 79.8 & 88.6 & 32.5 & 48.2 & 46.3 & 63.6 & 73.3 & 89.0 & 58.9 & 93.0 & 78.2 & 55.2 & 92.2 & 45.0 & 67.3 & 39.6 & 49.9 & 73.6 \\
\hline

PfD \cite{Richter2016} & 62.5 & 21.7 & 42.7 & 26.3 & 51.7 & 5.5 & 6.8 & 13.8 & 23.6 & 6.9 & 75.5 & 11.5 & 36.8 & 49.3 & 0.9 & 46.7 & 3.4 & 5.0 & 0.0 & 5.0 & 1.4 \\
CycleGAN \cite{Zhu2017} & 82.5 & 32.4 & 81.8 & 34.7 & 73.5 & 22.5 & 8.7 & 25.4 & 21.1 & 13.5 & 71.5 & 26.5 & 41.7 & 50.1 & 7.3 & 78.5 & 20.5 & 19.5 & 0.0 & 12.5 & 6.9 \\
CyCADA \cite{Hoffman2017} & - & 38.8 & 82.4 & 38.9 & 79.0 & 26.1 & 19.3 & 33.2 & 32.4 & 21.3 & 73.9 & 37.1 & 61.8 & 56.2 & 17.6 & 78.5 & 10.0 & 31.0 & 10.7 & 13.8 & 14.2 \\
UNIT \cite{Liu2017} & - & 36.1 & 79.2 & 28.5 & 75.9 & 22.1 & 13.6 & 27.0 & 29.7 & 18.8 & 75.9 & 25.8 & 56.3 & 57.5 & 21.8 & 81.1 & 18.9 & 21.6 & 1.5 & 13.7 & 17.2 \\
Our & - & \textbf{39.7} & \textbf{84.1} & 34.6 & \textbf{80.5} & 24.4 & 17.7 & 32.5 & 31.1 & \textbf{27.4} & \textbf{79.7} & 26.9 & \textbf{68.7} & \textbf{58.8} & 21.1 & \textbf{84.4} & \textbf{22.6} & 21.2 & 1.0 & 20.1 & \textbf{17.8} \\
\hline
\hline

CS \cite{Cordts2016} & 95.5 & 75.6 & 97.9 & 83.5 & 91.6 & 56.5 & 61.2 & 54.8 & 63.9 & 73.6 & 91.3 & 59.9 & 93.2 & 77.7 & 60.1 & 94.0 & 79.3 & 87.0 & 76.1 & 61.0 & 73.2\\
\hline

PfD \cite{Richter2016} & 82.9 & 40.0 & 79.2 & 26.9 & 79.5 & 19.1 & 27.4 & 13.8 & 23.6 & 6.9 & 75.5 & 11.5 & 36.8 & 49.3 & 0.9 & 46.7 & 3.4 & 5.0 & 0.0 & 5.0 & 1.4 \\
CycleGAN \cite{Zhu2017}	& 87.7 & 46.0 & 85.4 & 39.0 & 85.4 & 42.3 & 26.3 & 37.8 & 40.1 & 24.8 & 81.3 & 28.8 & 79.4 & 62.5 & 27.2 & 85.5 & 32.9 & 44.3 & 0.0 & 29.1 & 17.4 \\
CyCADA \cite{Hoffman2017} & 88.5 & \textbf{48.7} & 89.4 & 45.1 & 85.3 & 42.1 & 23.0 & 39.3 & 39.1 & 25.9 & 84.4 & 42.7 & 79.9 & 63.6 & 29.7 & 86.3 & 35.3 & 44.6 & 11.7 & 30.8 & 26.6\\
UNIT \cite{Liu2017} & 86.8 & 47.6 & 85.2 & 33.3 & 85.4 & 46.8 & 28.7 & 35.8 & 36.2 & 26.4 & 83.1 & 36.5 & 81.8 & 63.2 & 27.0 & 88.5 & 43.0 & 50.9 & 0.0 & 30.1 & 19.6\\
Our & \textbf{88.7} & 48.1 & \textbf{89.7} & 40.9 & \textbf{85.9} & 43.2 & 21.0 & 35.7 & 37.5 & \textbf{29.8} & 84.3 & 33.3 & \textbf{87.4} & 62.0 & 26.7 & 88.0 & \textbf{43.4} & \textbf{53.6} & 0.0 & 25.4 & 20.8\\
\hline
\hline

CS \cite{Cordts2016} & - & 70.8 & 97.3 & 79.5 & 90.1 & 40.1 & 50.7 & 51.3 & 56.1 & 67.0 & 90.6 & 59.0 & 92.9 & 76.7 & 54.2 & 92.9 & 68.8 & 80.6 & 68.5 & 58.0 & 71.7\\
\hline
ROAD \cite{Chen2018CVPR} & - & 35.9 & 85.4 & 31.2 & 78.6 & 27.9 & 22.2 & 21.9 & 23.7 & 11.4&  80.7 & 29.3 & 68.9 & 48.5 & 14.1 & 78.0 & 19.1 & 23.8 & 9.4 & 8.3 & 0.0\\
Adapt \cite{Tsai2018CVPR} & - & 41.4 & 86.5 & 25.9 & 79.8 & 22.1 & 20.0 & 23.6 & 33.1 & 21.8 & 81.8 & 25.9 & 75.9 & 57.3 & 26.2 & 76.3 & 29.8 & 32.1 & 7.2 & 29.5 & 32.5\\
Ours & - & \textbf{42.0} & 81.4 & 28.6 & \textbf{80.4} & 27.4 & 12.0 & \textbf{32.9} & \textbf{38.3} & \textbf{28.6} & \textbf{82.5} & 29.4 & \textbf{78.6} & \textbf{63.4} & 16.7 & \textbf{84.0} & 25.5 & \textbf{41.3} & 0.2 & 33.6 & 12.4\\
\end{tabular}}
%\end{center}
\caption{Semantic segmentation results for DRN26 (top), Deeplabv3 (mid) and Deeplabv2 (bottom) trained on translated images.}
\label{tab:drn}
\end{table*}

\section{Approach}

\subsection{Problem Definition}

Our setup consists of pairs of input images $x$ together with corresponding labels $y$ from synthetic dataset: $\{(x_i^s, y_i^s)\}_{i=1}^{N_s}$ and pairs $\{(x_j^r, y_j^r)\}_{j=1}^{N_r}$ from real. We denote input samples $\{x_i^s\}$ from the synthetic domain as $D_s$ and $\{x_j^r\}$ from real domain as $D_r$ and target domain as $D_s$: 

\begin{equation}
D_s=\{x_i^s\}_{i=1}^{N_s}
\end{equation}

\begin{equation}
D_r=\{x_j^r\}_{j=1}^{N_r}
\end{equation}

Let's consider variable $x$ in the input distribution space $\mathcal{X}$ taking values $x_i^s$, which are \textit{independent and identically distributed} and follows probability distribution $P_s(x)$:
\begin{equation}
\begin{split}
&x_i^s \in \mathcal{X}_s \subset \mathcal{X}, i = 0,1, ..., N_s\\
&\{x_i^s\}_{i=1}^{N_s} \sim P_s(x)
\end{split}
\end{equation}
The real samples $x_j^r$ in turn follow different probability distribution $P_r$:
\begin{equation}
\begin{split}
&x_i^r \in \mathcal{X}_r \subset \mathcal{X}, i = 0,1, ..., N_r\\
&\{x_j^r\}_{j=1}^{N_r} \sim P_r(x)
\end{split}
\end{equation}

In a sim-to-real setup both marginal distributions of $\{x^s_i\}$ and $\{x^t_j\}$ are generally different: $P_s(x) \neq P_r(x)$. This situation is addressed as a covariate shift \cite{Sugiyama2012} meaning that under the condition that $x$ is equivalent for both distributions, the conditional probability $P(y)$ is indistinguishable for $x^s$ and $x^r$.

Synthetic to real domain adaptation could be formalized as finding of a mapping function which translates samples from sub-space of synthetic domain into sub-space another $g: \mathcal{X}_s \rightarrow \mathcal{X}_r$.

Typically, such mapping function $g$ is approximated by a neural network, which training relies on adversarial loss (GAN) \cite{Goodfellow2014} in image space. During adversarial training one model called discriminator gets input samples from real distribution $P_r$ and from generative distribution $P(g(x^s))$. During this zero-sum game discriminator learns to distinguish real samples from synthesized by generator, which in turn learns to generate samples which are harder to distinguish. When training converges $g$ imposes real distribution on transformed samples $P(g(x^s)) = P_r$.

Adversarial loss applied in the image space works very well in making generated images similar to target ones. Thus, discriminator learns regularities in the real domain and imposes perturbations in the generated images. This not only makes generated images target-alike w.r.t. appearance but also introduces mismatch in content and semantic layout.

\subsection{Importance function}

To reduce semantic inconsistency in transferred samples we intend to correct the distribution bias between synthetic and real datasets. To achieve that and mitigate the impact of covariate shift on the learning procedure we employ the \textit{importance weighting} concept. The key idea of importance weighting is to consider informative training samples based on their \textit{importance}. Given density functions of both synthetic and real distributions, \textit{importance function} could be defined as:
\begin{equation}
\omega(x) = \frac{p_r(x)}{p_s(x)}
\end{equation}

\subsection{Density ratio estimation}
In unsupervised synthetic-to-real image transfer it is hard to estimate probability densities both for the source domain and the real domain without prior information about distributions. This could be however avoided when addressed \textit{density ratio estimation} directly. We rely in our approach on estimation technique called Kullback-Leibler importance estimation procedure (KLIEP), which has been introduced in \cite{Sugiyama2007}. This procedure focuses directly on \textit{density ratio estimation} between source and target densities instead of estimating them separately.

KLIEP aims to model the importance function $\omega(x)$ as:

\begin{equation}
\hat{\omega}(x) = \sum_l{\alpha_l\varphi_l(x)},
\end{equation}
where parameters ${\alpha_l}$ are supposed to be learned from samples ${x_i^s}$ (source) and ${x_j^t}$ (target) and ${\varphi_l(x)}$ are the basis functions. The estimation model $\hat{\omega}(x)$ approximates the target density: $\hat{p}_t(x) = \hat{\omega}(x)p_s(x)$. Parameters $\alpha_l$ of the model should be calculated in a way that Kullback-Leibler divergence from $p_t(x)$ to $\hat{p}_t(x)$ is minimized.
\begin{equation}
\begin{split}
&KL(p_t||\hat{p}_t) = \Exp_{x^t}\big[\log\frac{p_t(x)}{\hat{\omega}(x)p_s(x)}\big] \\ 
&= \Exp_{x^t}\big[\log\frac{p_t(x)}{p_s(x)}\big] - \Exp_{x^t}[\log\hat{\omega}(x)]\\
%&= \sum\limits_{j}^{N_r}{\log\frac{p_r(x_j^r)}{p_s(x_j^r)}}
\end{split}
\end{equation}
Since the first term does not depend on $\alpha$ we consider only the latter one:

\begin{equation}
\Exp_{x^t}[\log\hat{\omega}(x)] = \frac{1}{N_t}\sum\limits_{j}^{N_t}{\log{\sum\limits_{l}{\alpha_l\phi_l(x_j^t)}}}
\label{eq:j}
\end{equation}
Thus, in order to minimize KL divergence we can maximize the (\ref{eq:j}) w.r.t. $\alpha$ under following constraint:
\begin{equation}
\Exp_{x^s}[\hat{\omega}(x)] = \frac{1}{N_s}\sum\limits_{i}^{N_s}{{\sum\limits_{l}{\alpha_l\phi_l(x_i^s)}}} = 1,
\end{equation}

This constraint comes from the fact that $\hat{p_t}(x)$ is a probability density function itself. In that way we defined our optimization problem:

\begin{subequations}
\begin{alignat}{2}
&\mathop\text{maximize}_{\alpha_l} &\qquad& \sum\limits_{j}\log\sum\limits_{l}\alpha_l\varphi(x_j^t) \\
&\text{subject to} & & \sum\limits_l\alpha_l\sum\limits_i \varphi_l(x_i^s) = N_s,\\
& & & \alpha \geq 0.
\end{alignat}
\end{subequations}

We use RBF kernel $K_{\sigma_r}$ centered in target samples $x_j^t$ and width $\sigma_t$ (found by grid search maximizing \label{eq:J}) to calculate respective importance for source samples. Analogous calculations we perform reverse direction:

\begin{equation}
\begin{split}
\hat{\omega}(x^s) = \sum_l{\alpha_l K_{\sigma_r}(x^s, x_l^t)},\\
\hat{\psi}(x^t) = \sum_k{\beta_k K_{\sigma_s}(x^t, x_k^s)}
\end{split}
\end{equation}

We exploit gradient ascent with constraint satisfaction to find $\hat{\omega}(x^s)$ and $\hat{\psi}(x^t)$ in order to complete pre-matching of marginal distributions.

\subsection{Weighted loss}
In our adversarial domain adaptation approach we also rely on cycle-consistency loss, as it reveals robust training and produces consistent results in high resolution. Thus, our loss function is constructed by importance weighted adversarial losses \cite{Goodfellow2014} for both $g_r:\mathcal{X}_s \rarr \mathcal{X}_r$ and $g_s:\mathcal{X}_r \rarr \mathcal{X}_s$ and importance weighted cyclic-consistency losses \cite{Zhu2017}:
\begin{equation}
\begin{split}
&\Loss = \Loss_{IWAdv} + \Loss_{IWCyc}\\
& = \Exp_{x^r}[\psi(y^r)\log d_r(x^r)]\\
& + \Exp_{x^s}[\omega(y^s)\log(1 - d_r(g_r(x^s))]\\
& + \Exp_{x^s}[\omega(y^s)\log d_s(x^s)]\\
& + \Exp_{x^r}[\psi(y^r)\log(1 - d_s(g_s(x^r))]\\
& + \Exp_{x^r}[\psi(y^r)\|g_r(g_s(x^r)) - x^r\|]\\
& + \Exp_{x^s}[\omega(y^s)\|g_s(g_r(x^s)) - x^s\|]\\
\end{split}
\end{equation}

\section{Experiments}
To evaluate our approach we employ 2 experimental setups: toy example and real data. In the first one we simulate source and target distributions by Gaussian and uniform samplers. In the real setup we experiment with large scale datasets in simulated and real traffic scene environments.

\subsection{Toy Example}
\begin{figure}%[!t]
\includegraphics[width=1\columnwidth]{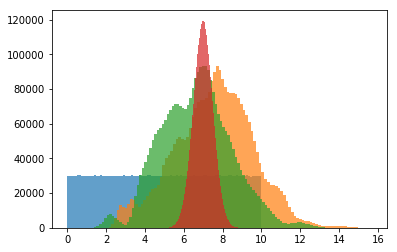}
\caption{Histograms for source data (blue), target data (red), generated by Vanilla GAN (orange), generated by our KLIEP GAN (green).}
\label{fig:toy_example}
\end{figure}

To facilitate the toy experiment we generate source and target datasets from uniform and normal distributions respectively. In this particular example both datasets consist of 10,000 random vectors of size 300. Those vectors of target dataset have been sampled from Gaussian distribution with mean value $7.0$ and standard deviation $0.5$, those of source dataset from uniform distribution in the segment $[0, 10)$. Histograms for both distributions could be observed in the figure~\ref{fig:toy_example} depicted with blue and red.

As a baseline for the toy example we train vanilla GAN model \cite{Goodfellow2014} on source and target dataset for 40 epochs with batch size of 200.The generator in the architecture of our choice consists of number of linear activations and scaled exponential linear units \cite{Klambauer2017}, while the discriminator net used additional sigmoid activation. We use SGD optimizer for the generator as well as for the discriminator with learning rates $8e-3$ and $4e-3$ respectively. The goal of the network is to transform input source distribution in a way that generated and target distributions are similar.

We extend the aforementioned vanilla GAN with the proposed KLIEP-based importance loss (\ref{eq:toy_loss}) and train this model following the same experimental setup as previously.

\begin{equation}
\begin{split}
&\Loss = \Loss_{IWAdv} = \Exp_{x^s}[\log d(x^t)]\\
& + \Exp_{x^s}[\sum_l{\alpha_l K_{\sigma_t}(x^s, x_l^t)}\log(1 - d(g(x^s))]\\
\end{split}
\label{eq:toy_loss}
\end{equation}
In both trainings we intentionally switch off random batches.

Both trained models, the vanilla GAN and KLIEP GAN one, were deployed on 10000 source vectors for inference. Resulting distributions were compared in terms of moments and distance to the target. They are depicted in the figure~\ref{fig:toy_example} in orange and green respectively. We evaluate generated distributions using Wasserstein and Energy distances between them and target distribution. Obtained results are reported in the table~\ref{tab:toy_results}.\\
From the results of the ablation study on the toy data we can tell that usage of the density ratio estimator for distribution pre-matching significantly improves the results of adversarial learning. Distribution generated with the importance loss is closer to target one in terms of the moments as well as in terms of distances. Wasserstein distance to the target distribution improves by 20\% and energy distance by 15\%.

%\begin{center}
\begin{table}%[!h]
\renewcommand{\arraystretch}{1.5}
\resizebox{\columnwidth}{!}{
\begin{tabular}{l|cc|cc}
\hline
\rotatebox[origin=l]{0}{ Distribution }
& \rotatebox[origin=c]{0}{ $\mu$ }
& \rotatebox[origin=c]{0}{ $\sigma$ }
& \rotatebox[origin=c]{0}{ Wasserstein distance}
& \rotatebox[origin=c]{0}{ Energy distance}
\\
\hline
Target (Gauss)	& 7.0 & 0.5 & - & - \\
\hline
Source (uniform)& 5.0 & 2.9 & 2.56 & 1.39 \\
Vanilla GAN		& 7.7 & 2.0 & 1.32 & 0.79 \\
Ours			& 6.7 & 1.8 & 1.08 & 0.67 \\
\hline
\end{tabular}
}
\caption{Distances between generated and target distributions (less is better).}
\label{tab:toy_results}
\end{table}
%\end{center}
\begin{figure*}[!h]

\begin{minipage}{0.16\textwidth}
\includegraphics[width=1\textwidth]{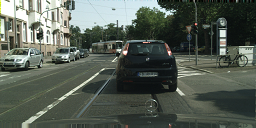}
\includegraphics[width=1\textwidth]{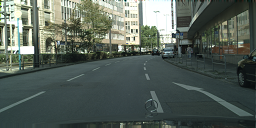}
\includegraphics[width=1\textwidth]{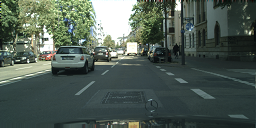}
\includegraphics[width=1\textwidth]{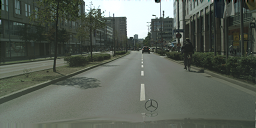}
\centering{Cityscapes}
\end{minipage}
\begin{minipage}{0.16\textwidth}
\includegraphics[width=1\textwidth]{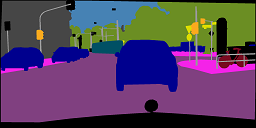}
\includegraphics[width=1\textwidth]{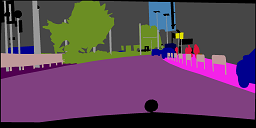}
\includegraphics[width=1\textwidth]{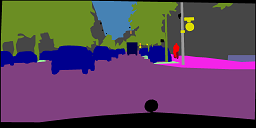}
\includegraphics[width=1\textwidth]{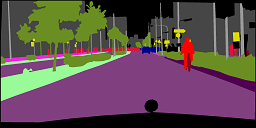}
\centering{Ground truth}
\end{minipage}
\begin{minipage}{0.16\textwidth}
\includegraphics[width=1\textwidth]{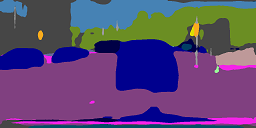}
\includegraphics[width=1\textwidth]{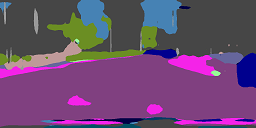}
\includegraphics[width=1\textwidth]{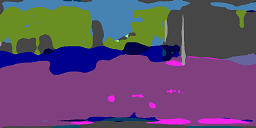}
\includegraphics[width=1\textwidth]{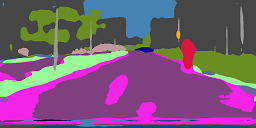}
\centering{CycleGAN}
\end{minipage}
\begin{minipage}{0.16\textwidth}
\includegraphics[width=1\textwidth]{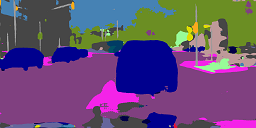}
\includegraphics[width=1\textwidth]{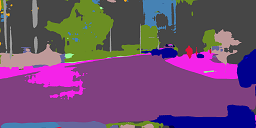}
\includegraphics[width=1\textwidth]{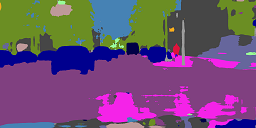}
\includegraphics[width=1\textwidth]{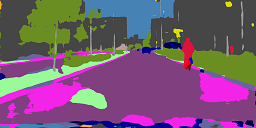}
\centering{CyCADA}
\end{minipage}
\begin{minipage}{0.16\textwidth}
\includegraphics[width=1\textwidth]{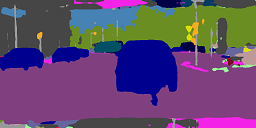}
\includegraphics[width=1\textwidth]{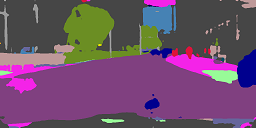}
\includegraphics[width=1\textwidth]{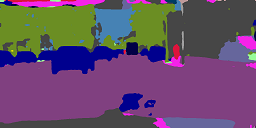}
\includegraphics[width=1\textwidth]{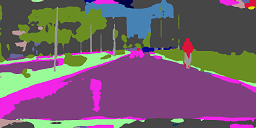}
\centering{UNIT}
\end{minipage}
\begin{minipage}{0.16\textwidth}
\includegraphics[width=1\textwidth]{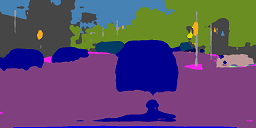}
\includegraphics[width=1\textwidth]{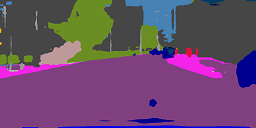}
\includegraphics[width=1\textwidth]{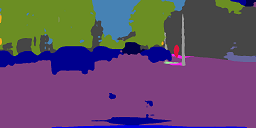}
\includegraphics[width=1\textwidth]{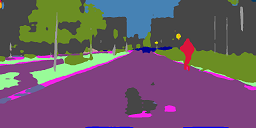}
\centering{Ours}
\end{minipage}

\caption{Examples of semantic segmentation by DRN26 trained on translated synthetic images.}
\label{fig:visual_results_drn_gta_cityscapes}
\end{figure*}
\begin{figure*}[!h]

\begin{minipage}{0.16\textwidth}
\includegraphics[width=1\textwidth]{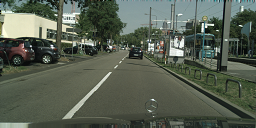}
\includegraphics[width=1\textwidth]{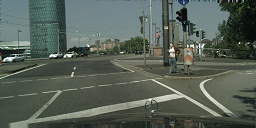}
\includegraphics[width=1\textwidth]{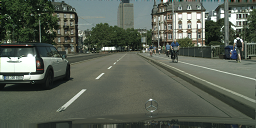}
\includegraphics[width=1\textwidth]{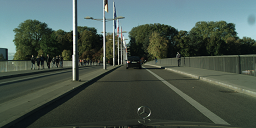}
\centering{Cityscapes}
\end{minipage}
\begin{minipage}{0.16\textwidth}
\includegraphics[width=1\textwidth]{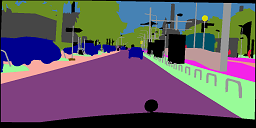}
\includegraphics[width=1\textwidth]{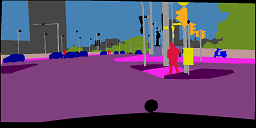}
\includegraphics[width=1\textwidth]{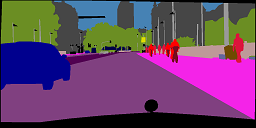}
\includegraphics[width=1\textwidth]{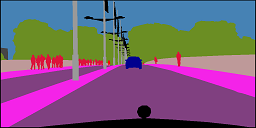}
\centering{Ground truth}
\end{minipage}
\begin{minipage}{0.16\textwidth}
\includegraphics[width=1\textwidth]{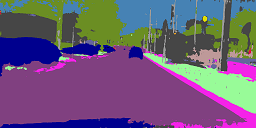}
\includegraphics[width=1\textwidth]{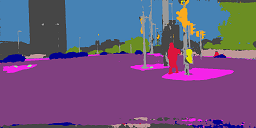}
\includegraphics[width=1\textwidth]{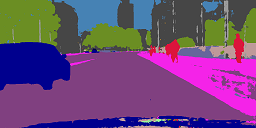}
\includegraphics[width=1\textwidth]{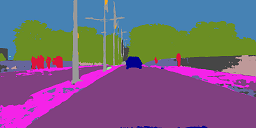}
\centering{CycleGAN}
\end{minipage}
\begin{minipage}{0.16\textwidth}
\includegraphics[width=1\textwidth]{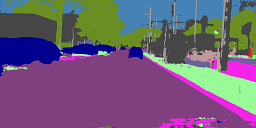}
\includegraphics[width=1\textwidth]{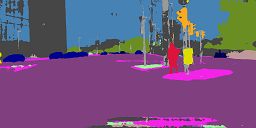}
\includegraphics[width=1\textwidth]{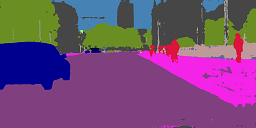}
\includegraphics[width=1\textwidth]{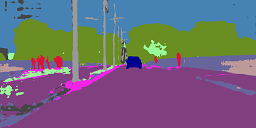}
\centering{CyCADA}
\end{minipage}
\begin{minipage}{0.16\textwidth}
\includegraphics[width=1\textwidth]{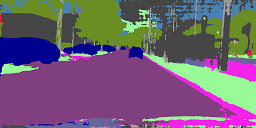}
\includegraphics[width=1\textwidth]{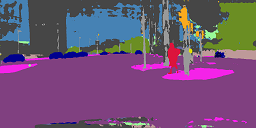}
\includegraphics[width=1\textwidth]{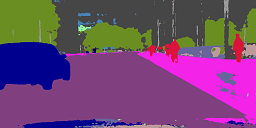}
\includegraphics[width=1\textwidth]{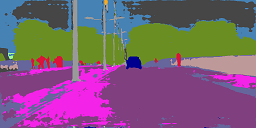}
\centering{UNIT}
\end{minipage}
\begin{minipage}{0.16\textwidth}
\includegraphics[width=1\textwidth]{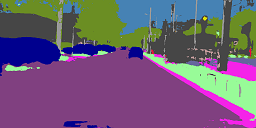}
\includegraphics[width=1\textwidth]{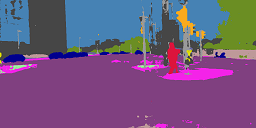}
\includegraphics[width=1\textwidth]{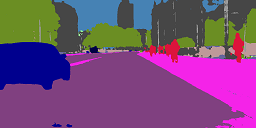}
\includegraphics[width=1\textwidth]{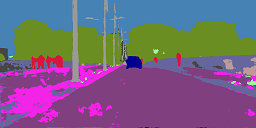}
\centering{Ours}
\end{minipage}

\caption{Examples of semantic segmentation by Deeplabv3 trained on translated synthetic images.}
\label{fig:visual_results_deeplab_gta_cityscapes}
\end{figure*}
\subsection{Real Data}
Similarly our large scale evaluation pipeline consists of 2 stages as well. First we train our domain adaptation network with synthetic and real datasets. Then we deploy it on the synthetic images and translate them to real domain. In the second stage we train multiple target prediction task models with translated images and evaluate theirs performance on real validation dataset.

As a real dataset we take one of the recent dataset called Cityscapes \cite{Cordts2016} as most commonly used in autonomous driving community. It provides 5000 frames of urban traffic scenes of resolution 2048$\times$1024 alongside with fine pixel-level semantic labels. Samples are split into training, validation and test subsets. It enfolds 50 cities multiple times of the year, different daylight and weather conditions. Cityscapes provides ground-truth for semantic, instance and pan-optic segmentation. Semantic segmentation covers 30 classes and also single instances annotations for dynamic objects such as \textit{car, person, rider etc}. We focus evaluation on 19 training classes: \textit{road, building, sky, sidewalk, vegetation, car, terrain, wall, truck, pole, fence, bus, person, traffic light, traffic sign, train, motorcycle, rider, bicycle}.

As synthetic one we utilize the dataset from \cite{Richter2016} as a most comprehensive synthetic dataset. It provides almost 25000 frames acquired from a computer game engine alongside with semantic labels. Every frame is of resolution 1914 $\times$ 1052. Although, it reveals some labeling bugs it remains the main synthetic dataset for autonomous driving. It shows certain advantages in comparison with other synthetic datasets w.r.t traffic scenes. It is by far more realistic in terms of appearance as well as in terms of traffic scene construction. It shows a huge variance in scenery, scenarios and appearance.

\begin{figure*}%[!h]
\begin{minipage}{0.195\textwidth}
\includegraphics[width=1\textwidth]{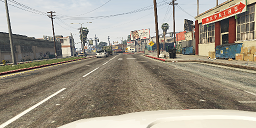}
\centering{Original}
\end{minipage}
\begin{minipage}{0.195\textwidth}
\includegraphics[width=1\textwidth]{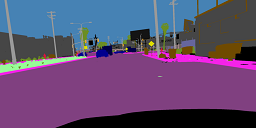}
\centering{Ground Truth}
\end{minipage}
\begin{minipage}{0.195\textwidth}
\includegraphics[width=1\textwidth]{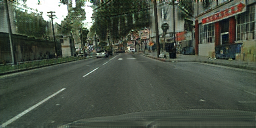}
\centering{Low}
\end{minipage}
\begin{minipage}{0.195\textwidth}
\includegraphics[width=1\textwidth]{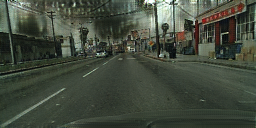}
\centering{Medium}
\end{minipage}
\begin{minipage}{0.195\textwidth}
\includegraphics[width=1\textwidth]{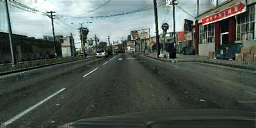}
\centering{High}
\end{minipage}

\caption{Examples of image transfer by KLIEP GAN trained on translated synthetic images of different importance cohorts.}
\label{fig:ablation}
\end{figure*}

First, we evaluate the results qualitatively. Results of domain adaptation in comparison with other approaches could be seen on the figure~\ref{fig:adaptation}. On this figure one can see that translation by multiple models introduces mismatching patches in place of the classes e.g. vegetation and sky. In turn density ratio prematching enables translation model to preserve semantics.

Most importantly, we evaluate quality of image transfer on the semantic segmentation task. For this evaluation we train state-of-the-art segmentation models on our generated data and evaluate on Cityscapes val dataset. Needs to be said that during the training segmentation model did not "see" any real images from target dataset.

We follow the original works in our evaluation experiments. As a preprocessing step all images were down-scaled to 1024 $\times$ 512 pixels resolution. In our evaluation we rely on DRN \cite{Yu2017} and Deeplabv3 \cite{Chen2018}. DRN26 was initialized on the weights pretrained on Imagenet \cite{Krizevsky2012} and fine-tuned for 200 Epochs with random crops 600 $\times$ 600 of our translated data with momentum 0.99 and learning rate 0.001 decreasing by 10 every 100. Deeplabv3 utilizes xception65 backbone an has been trained for 90,000 steps with batch size of 16, we keep learning rate of 0.007 and crops of 513 $\times$ 513. Obtained metrics for the best performing snapshots for both networks are reported in the table~\ref{tab:drn}. Additionally we train Deeplabv2 \cite{Chen14} and evaluate on Cityscapes \textit{val}. In this evaluation we also follow the setup of original work.

Our main metric is IoU or \textit{Jaccard Index} for particular class it calculates ratio of correctly classified pixels relatively to true positive, false positive and false negative predictions summed \cite{Everingham2015}. We additionally report its mean value over all 19 classes. This metric helps to take into consideration segmentation performance not affected by the size of particular class itself. We report however pixel accuracy as well. The results obtained in our experiments are presented
in the table~\ref{tab:drn}. The tables show performance of DRN26 and Deeplab networks trained on dataset generated by translation of synthetic images to real ones. Additionally we provide comparison numbers for the aforementioned nets on merely real (CS) and synthetic data (PfD).

In the table~\ref{tab:drn} one can see that pre-matching densities using KLIEP could improve performance by meanIoU and also by major classes such as \textit{road, building, vegetation, sky and car}. Class \textit{sky} has shown improvement by almost 7\% other classes by more than 2\%. For the Deeplab CyCADA remains top performing model w.r.t meanIoU but was improved by densities pre-matching for multiple classes as \textit{building, vegetation, sky, truck and bus}.

\subsection{Ablation Study}

Additionally to the toy example we perform ablation study also on the large scale datasets. The intention is to show how importance estimation in our KLIEP GAN influences the adversarial training. For that purpose we split the source dataset samples according to their importance estimates into 3 equal cohorts. Each cohort consists of 8322 training pairs and represents certain importance range: low, medium and high. Following evaluation steps greatly reproduce our main evaluation pipeline. We downscale though all samples to the resolution of 512$\times$256 to speed up the evaluation process. We train 3 instances of our KLIEP GAN model on the respective cohort as a source dataset and Cityscapes \textit{train} as a target dataset. After that, each of 3 models is deployed on \cite{Richter2016} dataset, which results in 3 adapted datasets with 24,966 transferred samples each. As a final step, we train 3 DRN models on the corresponding generated dataset and evaluate them on Cityscapes \textit{val}.

The results for of the ablation study are reported in the table~\ref{tab:ablation}. Here one can see the IoU values provided for major classes as well as meanIoU for all 19 original classes. The numbers reported in the ablation study confirm the intuition that the higher importances estimated by KLIEP reflect similarity with the target distributions. Thus, one can say that learning from more informative (with higher importance score) samples improves quality of adversarial image translation. Such qualitative improvement could be observed in figure~\ref{fig:ablation}. Table~\ref{tab:ablation} also confirms gradual improvement of translation quality as we move from low importance cohort to high importance meanIoU raises.

\begin{table}%[!h]
%\begin{center}
\renewcommand{\arraystretch}{1.5}
\resizebox{\columnwidth}{!}{
\begin{tabular}{l|c|cccccccc}
\hline
\rotatebox[origin=c]{90}{ Method }
& \rotatebox[origin=c]{90}{ mean IoU }
& \rotatebox[origin=r]{90}{ road }
& \rotatebox[origin=c]{90}{ sidewalk }
& \rotatebox[origin=c]{90}{ building }
& \rotatebox[origin=c]{90}{ wall }
& \rotatebox[origin=c]{90}{ fence }
& \rotatebox[origin=c]{90}{ vegetation }
& \rotatebox[origin=c]{90}{ sky }
& \rotatebox[origin=c]{90}{ car }
\\
\hline
Cityscapes \cite{Cordts2016}& 67.4 & 97.3 & 79.8 & 88.6 & 32.5 & 48.2 & 89.0 & 93.0 & 92.2 \\
\hline
No adapt \cite{Richter2016}	& 21.7 & 42.7 & 26.3 & 51.7 &  5.5 & 6.8 & 75.5 & 36.8 & 46.7 \\
\hline
Low 	& 27.9 & 75.6 & 28.7 & 69.1 & 14.5 & 18.5 & 63.4 & 45.8 & 75.7 \\
Medium 	& 28.3 & 77.8 & 24.0 & 71.6 & 10.7 & 17.1 & 69.3 & 69.6 & 73.5 \\
High 	& 30.2 & 82.2 & 40.2 & 72.1 & 15.3 & 23.2 & 72.9 & 69.5 & 77.6 \\
\hline
\end{tabular}}
%\end{center}
\caption{IoU values for semantic segmentation prediction by DRN26 trained on translated synthetic to real images obtained from different importance cohorts.}
\label{tab:ablation}
\end{table}

\section{Conclusion}
In this paper we proposed the usage of the density prematching domain adaptation based on KLIEP density ratio estimation procedure combined with effective cycle-consistency loss in order to tackle class covariate shift problem in synthetic and real datasets. We have shown in our experiments that this strategy works well for synthetic to real domain adaptation. First, we visualized the effects of KLIEP based loss of our model on the toy example. Here we have shown that distribution pre-matching is very helpful mean by adversarial learning of target distribution. In our large scale experiment we have shown that KLIEP loss not only improves visual quality of transferred synthetic to real image (mainly in terms of semantical consistency) but also improves performance of deep semantic segmentation network trained on the translated images (improvement by highly imbalanced classes such as vegetation and sky achieved $>$7\%). And finally our ablation study visualized how importance scores obtained by KLIEP affect adversarial training of the model.

\section{Acknowledgement}
The research leading to these results is funded by the German Federal Ministry for Economic Affairs and Energy within the project “KI Absicherung – Safe AI for Automated Driving". The authors would like to thank the consortium for the successful cooperation.

\newpage
{
\bibliographystyle{ieee}
\bibliography{literature}
}

\end{document}